\documentclass[final,3p,times]{elsarticle}
\usepackage{algorithmic,algorithm}
\usepackage{amssymb}
\usepackage{bm}
\usepackage{color}
\usepackage{amsmath}
\usepackage[tight,footnotesize]{subfigure}
\usepackage{multirow}

\newcommand{\ie}{{\it i.e.}}

\begin{document}

\begin{frontmatter}

\title{Active Multi-Kernel Domain Adaptation for Hyperspectral Image Classification}

\author[xidian]{Cheng Deng}
\author[beihang]{Xianglong Liu\corref{mycorrespondingauthor}}
\cortext[mycorrespondingauthor]{Corresponding author}
\ead{xlliu@nlsde.buaa.edu.com}
\author[xidian]{Chao Li}
\author[aus]{Dacheng Tao}
\address[xidian]{School of Electronic Engineering, Xidian University, Xi'an 710071, Shaanxi, China}
\address[beihang]{State Key Lab of Software Development Environment, Beihang University, Beijing 100191, China}
\address[aus]{Centre for Quantum Computation and Intelligent Systems, Faculty of Engineering and Information Technology, University of Technology, Sydney, PO Box 123, Broadway NSW 2007, Australia}

\begin{abstract}
Recent years have witnessed the quick progress of the hyperspectral images (HSI) classification. Most of existing studies either heavily rely on the expensive label information using the supervised learning or can hardly exploit the discriminative information borrowed from related domains. To address this issues, in this paper we show a novel framework addressing HSI classification based on the domain adaptation (DA) with active learning (AL). The main idea of our method is to retrain the multi-kernel classifier by utilizing the available labeled samples from source domain, and adding minimum number of the most informative samples with active queries in the target domain. The proposed method adaptively combines multiple kernels, forming a DA classifier that minimizes the bias between the source and target domains. Further equipped with the nested actively updating process, it sequentially expands the training set and gradually converges to a satisfying level of classification performance. We study this active adaptation framework with the Margin Sampling (MS) strategy in the HSI classification task. Our experimental results on two popular HSI datasets demonstrate its effectiveness.
\end{abstract}


\end{frontmatter}

\section{Introduction}
A very challenging problem in the remote sensing community is to generate land-cover maps for semantically characterizing Earth's surface \cite{DBLP:journals/lgrs/AlajlanPMF14,Li:2015,Li:2016,Tong:2016,Tang:2016,Li:2017}. {As one of the most widely used approaches, hyperspectral image (HSI) classification has recently gained in popularity and attracted research interests from other scientific disciplines such as image processing, machine learning, and computer vision \cite{Plaza:2009,Tarabalka:2012,Li:2014,Samat:2016,Ye:2016,Yan:2016,Zhang:2016,Shao:2017,Appice:2017}. Most of these studies belong to the supervised learning methods, which have shown promising classification performance in practice. However, they usually require many labeled samples to train classifier's parameters properly, which is quite expensive and time-consuming for real-world applications. Moreover, the high dimensionality of HSI makes it difficult to find an expected classifier only with a few labeled samples~\cite{Ifarraguerri:2000,DBLP:journals/tgrs/RajanGC08}.}

To address these issues, one feasible way is to exploit the available information from other geographical areas with abundant labeled samples (regarded as source domain). However, usually these areas are different from the target one, and there always exist certain shifts in data distribution, especially for the image data with underlying structures \cite{Chen:2015}. From a perspective of machine learning, this shift problem can be modeled by transfer learning, especially the domain adaptation (DA) approaches. In such scenario, it is always assumed that source domain and target domain possess similar characteristics, i.e., they share the same set of label classes or correlated class distributions \cite{DBLP:journals/pami/BruzzoneM10}.

A number of DA techniques have been adopted in HSI classification tasks~\cite{DBLP:journals/pami/BruzzoneM10, zhuo, DBLP:journals/tgrs/BruzzoneF01}, where both labeled samples from source domain and unlabeled samples from target domain are exploited to train a classifier for the target domain. \cite{DBLP:journals/pami/BruzzoneM10} proposed a DA framework named DASVM, which extended Transductive SVM (T-SVM) to label unlabeled target samples and remove some source samples progressively. In \cite{zhuo}, a multiple-kernel DA technique was designed to learn a discriminative model by simultaneously minimizing both the SVM structural risk function and the distribution mismatch. The DA work of~\cite{DBLP:journals/tgrs/BruzzoneF01} adapted the maximum-likelihood classifier to target domain through updating the classifier parameters. 

Most of the existing domain adaptation works assume that labeled samples are available only for the source domain, but not for the target domain, or only a few labeled data exists in target domain. In fact, the labeled information in target domain can directly and notably improve the classification performance. To alleviate the expensive cost on the collection of labeled data, an effective solution for discriminative classifier training is to interactively generate the labeled data using the active learning (AL) technique. In the AL literature, we know that some samples from the target domain can be selected and further labeled by the user, which finally form a new training set together with the existing training set in the source domain in order to adapt the classifier to the target domain \cite{DBLP:conf/igarss/PerselloB11}. There are a number of studies that have already taken the advantages of the AL strategy in HSI classification \cite{5764734,DBLP:journals/tgrs/PerselloB12,DBLP:journals/tgrs/PerselloB12,Dai:2007:BTL:1273496.1273521,Tuia2011}. However, they usually either only focus on the target domain without exploiting the useful information of the source domain, or lack of the power of capturing the data set shift using the ineffective active queries selection without fully utilizing the domain correlations.

{In this paper, we first propose a novel framework named multi-kernel learning with active learning (MKL-AL for short) for HSI classification, which combines the powerful AL and DA techniques based on the multiple kernels and largely helps compensate the data shift occurred between two image acquisitions. The main idea of our method is to retrain a multi-kernel classifier using the labeled samples from both source domain and the user-labeled samples selected from target domain, and the process of retraining is convergent to a satisfying performance of the desired level. On one side, the AL technique can enable us to fully utilize the target information. On the other side, DA based on multi-kernel learning can offer a good distribution distance measurement across domains, which helps us determine which kernel space is most fit for the data of the two domains and thus select the informative samples.} To illustrate the performance of our MKL-AL framework, here we choose margin sampling (MS) \cite{schohn2000less}\cite{DBLP:journals/prl/MitraSP04}\cite{DBLP:journals/tgrs/DemirPB11}, simple and common used uncertainty criterion, as the active learning strategy \cite{DBLP:journals/spm/Camps-VallsTBB14}\cite{DBLP:journals/pieee/CrawfordTY13}\cite{DBLP:journals/tgrs/DemirPB11} for selecting most informative pixels. Therefore, in the following, we can specialize our proposed MKL-AL method as multi-kernel learning with margin sampling (MKL-MS for short). We conducted extensive experiments on HSI classification over two hyperspectral datasets, and the experimental results demonstrate the effectiveness of our proposed framework.

The rest of this paper is organized as follows. The following section introduces the related works. Section \ref{sec:frame} presents our framework for hyperspectral image classification and elaborates on the detailed components of the proposed framework including the multi-kernel learning, domain adaptation and the active learning MS strategy. In Section \ref{sec:exp}, we present and discuss experimental results. Finally, we conclude in Section \ref{sec:con}.

\section{Related Works}
{In the literature, support vector machine (SVM) has become one of the most successfully used techniques for hyperspectral image classification \cite{Melgani:2004}, mainly due to the fact that SVM can deal with the high-dimensional and noisy data, with the help of the sparse set of the support vectors \cite{Xu:2017} and the powerful kernel tricks \cite{Liu:2014:pr}. This has been proved in many applications like biological problems, which involve high-dimensional, noisy data, for which SVMs are known to behave well compared to other statistical or machine learning methods \cite{Bandyopadhyay:2007}}. Therefore, many kernel-based SVM methods have been studied to capture the complex semantic structure of the hyperspectal images. Basically, these methods first uplift data in the original feature space to a high-dimensional kernel space, and then solve the linear classification problem in the uplifted space. With the supervised information from many labeled samples fed to the model, the kernel-based solutions have demonstrated excellent performance in hyperspectral data classification, in terms of accuracy and robustness \cite{Camps:2005,Camps:2006,Camps:2007,Liu:2016,Li:2016}.

Most of these existing HSI classification methods assume that labeled samples are available for the concerned domains. However, in practice it is common that there exist different domains in the hyperspectral images and some of them do not have sufficient labeled samples, because it is usually expensive to collect labels for all domains. Therefore, respectively training a classifier for each domain becomes infeasible. At the same time, due to the spectral shifts among the domains, the model trained in one domain cannot directly fit the other domains. This will obviously limit the power of the traditional HSI classification methods. To address the problem, the domain adaptation (DA) serves as a successful strategy that transfers the well trained classifier from a source domain to the different, yet related target domain \cite{Tuia:2016}. In DA based HSI classification, both labeled samples from the source domain and unlabeled samples from the target domain are exploited to train a classifier for the target domain. Such a technique has been proved able to avoid the expensive and time-consuming labelling efforts, and meanwhile achieve satisfying classification performance across domains \cite{DBLP:journals/pami/BruzzoneM10, zhuo, DBLP:journals/tgrs/BruzzoneF01}.

There are several ways in DA research to migrate the classifiers among the source and the target domains. One of the typical strategies is to make the data distributions more similar across the domains to train a single model that can simultaneously classify the source and target domains. \cite{zhuo} designed a multiple-kernel DA technique to learn a discriminative model by simultaneously minimizing both the SVM structural risk and the distribution divergence. \cite{DBLP:journals/tgrs/BruzzoneF01} adapted the maximum-likelihood classifier to the target domain through updating the classifier parameters. Even with the promising progress achieved by the DA based HSI classification methods, it is still beyond the desired performance without the supervised information from target domain, mainly due to the divergence among the source and the target domains.

{Since the labeled information in target domain can directly and notably improve the classification performance, it is obviously helpful that we can exploit a small number of labeled data to boost the classification performance, which at the same time only brings a quite limited additional cost for the labeled data collection. The active learning (AL) strategies have been widely studied in the literature to tackle such a challenging task in recent years \cite{Tong:2002,Jain:2010,Huang:2014,Liu:2016:cvpr}, with the aim to exploit the information available from unlabeled data and to enrich the labeled data. In AL process, labelling originally unlabeled data is usually completed by a user according to an specific informative measure.}

Due to the promising performance, many efforts have been devoted to the integration of AL in domain adaptation based HSI classification \cite{DBLP:conf/igarss/PerselloB11,DBLP:journals/tgrs/PerselloB12,Tuia2011,6353565,5764734}. \cite{DBLP:journals/tgrs/PerselloB12} offered an iterative AL process by adding the most informative samples to the training set, while removing the source-domain samples that do not fit with the distributions of the classes in the target domain. \cite{6353565} is a framework that efficiently combines the DA and AL techniques, and the most informative pixels are sampled with active queries from the target image while adapting the obtained classifier using a transfer learning strategy. In \cite{Tuia2011}, a DA framework equipped with the active selection pursued the training samples in unknown areas using the strategies based on uncertainty and clustering. These active learning methods are able to alleviate the expensive cost on the collection of labeled data by interactively generating the labeled data.

\section{Active Multi-Kernel Domain Adaptation}\label{sec:frame}
First of all, we introduce the notations adopted throughout this paper. Let $D_S = \mathbf{x}_i^S|_{i=1}^{N_S}$ be the $N_S$ labeled samples (\ie, pixels) in source domain with the corresponding labels $y_i^S|_{i=1}^{N_S}$, $\mathcal{D}_T = \mathbf{x}_j^T|_{j=1}^{N_T}$ be the $N_T$ unlabeled samples in target domain, and $\mathcal{D}_L = \mathcal{D}_S \cup \mathcal{D}_C$, $\mathcal{D}_C=\left\{\mathbf{x}_i | \mathbf{x}_i\in \mathcal{D}_T \textit{ and labeled in the active learning}\right\}$ be all the labeled samples from both source and target domains, which will be iteratively updated in our active multi-kernel domain adaptation. Here, we adopt binary classifier for simplicity and easy understanding, so each label $y_i\in \{-1, +1\}$. In practical scenarios, this can be further extended using the one-against-all strategy for the multi-class problem.

\begin{figure*}[!t]
\centering
\includegraphics[width=0.8\linewidth]{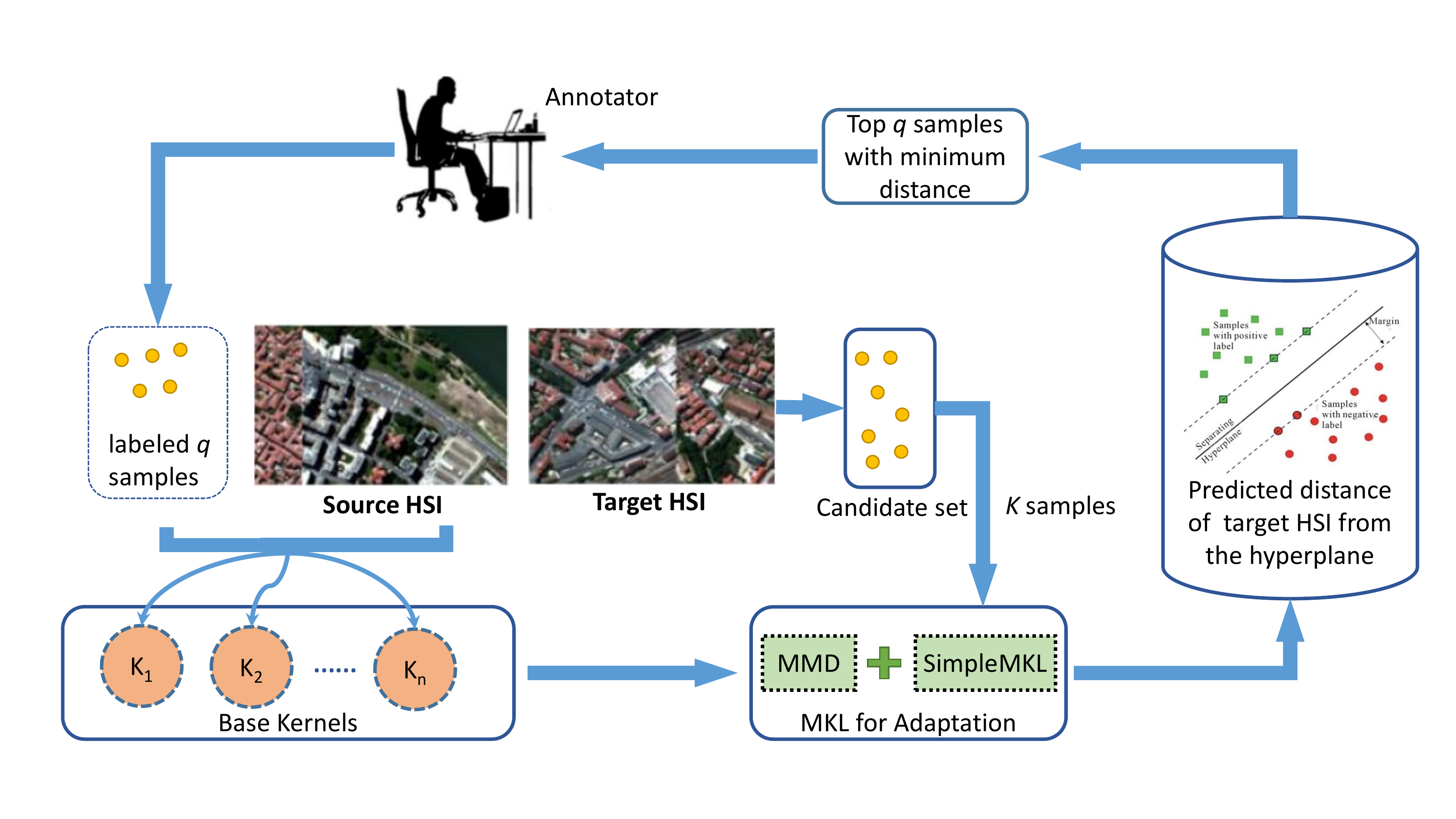}
\caption{{The flowchart of our proposed MKL-AL method. It consists of two parts: the active learning for target data labelling (the top part) and  multi-kernel based domain adaptation (the bottom part). The active learning technique heuristically updates the labeled target domain according to the MS selection criteria, while the domain adaptation is applied to the retraining of multi-kernel classifier based on the adaption from source domains to the target.}}
\label{flowchart}
\end{figure*}

{The flowchart in Fig. \ref{flowchart} outlines the general procedure, including the updating of the base kernels and the maximum mean discrepancy. It consists of two parts, of which the first one is the active learning for target data labelling, with MS selection criteria that heuristically updates the labeled dataset from both source and target domain (the top part), and the other corresponds to the retraining of multi-kernel classifier based on the adaption from source domains to the target (the bottom part). In active learning, the most interesting candidates for labelling are the ones that fall within the margin of the current classifier, as they are the most likely to become new support vectors \cite{devis1}. So the most interesting candidates from the target domain are identified by using the margin sampling (MS) strategy. After assigning the corresponding true labels by annotators, these candidates are further added to training data $\mathcal{D}_L$ in target domain for a better MKL classifier.}

\subsection{Multi-Kernel Learning for HSI Classification}\label{sec:mkl}
The multiple kernel learning (MKL) framework has been proved powerful for support vector machine (SVM) based classification in the literature \cite{Vishwanathan:2010}. Therefore, in this paper we employ MKL technique for HSI classification. Specifically, we pursue a decision function of the form $f(\mathbf{x})+b = \sum_m f_m(\mathbf{x})+b$, where each function $f_m$ belongs to a different reproducing kernel Hilbert space (RKHS). According to the above functional framework, the well-known SimpleMKL method proposed a weighted 2-norm regularization formulation with an constraint on the combination weights, which encourage the sparsity \cite{simplemkl}. {Based on the combination, it solves a standard SVM optimization problem with a kernel defined as a linear combination of multiple kernels. Supposing $\mathcal{D}_L$ contains $N_L$ labeled samples (or pixels), the MKL based SVM problem can be addressed by solving the following convex problem}, which we will be referred to as the primal MKL problem in SimpleMKL
\begin{equation}
\begin{split}
\min\limits_{f_m,b,\mathbf{d},\boldsymbol{\xi}} \ & \frac{1}{2}\sum_{m=1}^{M}\frac{\|f_m\|_{\emph{H}_m}^2}{d_m}+C\sum_{i=1}^{N_L}\xi_i \\
{s.t.}\  &  y_i \left(\sum_{m=1}^{M} f_m(\mathbf{x}_i)+b\right)\geq 1-\xi_i, \ \xi_i\geq 0, \ \mathbf{x}_i \in \mathcal{D}_L\\
 &\mathbf{d}'\mathbf{1}=1, \ {d}_m\geq\mathbf{0}
\end{split}
\end{equation}
where $\mathbf{d}$ is the linear combination coefficients for $M$ components, and each $d_m$ controls the contribution of each component $f_m$ in the objective function. The smaller $d_m$ means the smoother $f_m$, according to the measurement $\|f_m\|_{\emph{H}_m}$ (Note that when $d_m = 0$, $\|f_m\|_{\emph{H}_m}$ has also to be set to zero for a finite objective value). The constraint on $\mathbf{d}$ actually corresponds to a the popular sparsity $\ell_1$-norm constraint, which forces some $d_m$ to be zero and thus encourages sparse basis kernel expansions. Note that since the above formulation is convex and differentiable, it is easy to solve the problem by a simple gradient method \cite{simplemkl}. 

In SVM based classification, we usually adopt a linear projection based classifier. Specifically, we use the form $f_m(\mathbf{x}) = \mathbf{w}_m' \phi_m(\mathbf{x})$, where $\mathbf{w}_m$ is the projection vectors, and  $\phi_m(\cdot)$ is the nonlinear feature mapping function, which induces the kernel function $k_m(\cdot, \cdot)$, \ie, $k_m(\mathbf{x}_i, \mathbf{x}_j)=\phi_m(\mathbf{x}_i)'\phi_m(\mathbf{x}_j)$. Based on the linear classifier formulation, the above MKL problem can be further rewritten as follows
\begin{equation}
\begin{split}
\min\limits_{\mathbf{w}_m,b,\mathbf{d},\boldsymbol{\xi}} \ & \frac{1}{2}\sum_{m=1}^{M}\frac{\|\mathbf{w}_m\|^2}{d_m}+C\sum_{i=1}^{N_L}\xi_i \\
{s.t.}\  &  y_i \left(\sum_{m=1}^{M} \mathbf{w}_m'\phi_m(\mathbf{x}_i)+b\right)\geq 1-\xi_i, \ \xi_i\geq 0, \ \mathbf{x}_i \in \mathcal{D}_L\\
 &\mathbf{d}'\mathbf{1}=1, \ {d}_m\geq\mathbf{0}
\end{split}
\end{equation}

Such a formulation is quite similar to the standard SVM, \ie, given the combination weight $\mathbf{d}$, its dual problem can be easily obtained using the Lagrangian multipliers satisfying KKT conditions. According to the theoretical result in \cite{simplemkl}, the above optimization problem can be turn to its associated dual problem as follows:
\begin{equation}\label{svm}
G(\mathbf{d}) =\left\{
\begin{aligned}
 \max\limits_{\boldsymbol{\alpha}} \ & \boldsymbol{\alpha}'\mathbf{1}-\frac{1}{2}(\boldsymbol{\alpha}\circ\mathbf{y})'\left(\sum_{m=1}^{M}d_m \mathbf{K}_m^{L,L}\right)(\boldsymbol{\alpha}\circ\mathbf{y}) \\
{s.t.}\  &  \boldsymbol{\alpha}'\mathbf{y} =0 \ , 0 \leq \alpha_i \leq C
\end{aligned}
\right.
\end{equation}
where $\mathbf{K}_m^{L,L}\in \mathbb{R}^{N_L\times N_L}$ is the kernel matrix defined for the labeled data by $k_m(\cdot,\cdot)$, and {$\circ$ is the Hadamard product operator which performs the product in an elementwise manner}.

\subsection{Domain Adaptation based on MKL}\label{sec:da}
In our MKL framework, besides the label information from the training samples, we further consider the correlation between the source and target domains for HSI classification, and incorporate the domain adaptation technique into the MKL based classification. Intuitively, since we have more training data from the source domain than those from the target domain, it is natural to find the optimal migration between the two domains. Therefore, we attempt to reduce the distribution discrepancy when we transfer from source domains to the target one.

To address this issue, we follow \cite{duan1} and develop the SimpleMKL formulation with a Maximum Mean Discrepancy (MMD). The MMD measures the mismatch based on the distance between the means of the samples, respectively, from the source domain and the target domain in a RKHS, namely,
\begin{equation}
\label{da}
\Omega = \sum_{m=1}^M \left\|\frac{1}{N_S}\sum_{\mathbf{x}_i \in \mathcal{D}_S}\phi_m(\mathbf{x}_i)- \frac{1}{N_T}\sum_{\mathbf{x}_j \in \mathcal{D}_T}\phi_m(\mathbf{x}_j)\right\|_{\emph{H}_m}^2.
\end{equation}
If we define the matrix $\mathbf{L}=(L_{ij})$:
\begin{equation}
L_{ij}=\left\{
\begin{aligned}
&\frac{1}{N_S^2},  & \text{if}\ \mathbf{x}_i,\ \mathbf{x}_j\in \mathcal{D}_S;\\
&\frac{1}{N_T^2},  & \text{if}\ \mathbf{x}_i,\ \mathbf{x}_j\in \mathcal{D}_T;\\
&-\frac{1}{N_SN_T}, & \text{otherwise}
\end{aligned}
\right.
\end{equation}
then $\Omega$ will turn to
\begin{equation}
\label{eq-2}
\Omega(\mathbf{d}) = \sum_{m=1}^M d_m \text{tr}(\mathbf{K}_m \mathbf{L}) = \mathbf{d}'\mathbf{k},
\end{equation}
where $\mathbf{k}=\left[\text{tr}(\mathbf{K}_1\mathbf{L}),\ldots, \text{tr}(\mathbf{K}_M\mathbf{L})\right]'$, with each kernel component induced from $\phi_m$ defined as
\begin{equation}
\label{kernel}
\mathbf{K}_m=\begin{pmatrix}\mathbf{K}_m^{S,S}&\mathbf{K}_m^{S,T} \\ \mathbf{K}_m^{T,S}&\mathbf{K}_m^{T,T}\\ \end{pmatrix}\in \mathbb{R}^{(N_S+N_T)\times(N_S+N_T)}
\end{equation}
where $\mathbf{K}_m^{S,S}\in \mathbb{R}^{N_S\times N_S}$, $\mathbf{K}_m^{T,T}\in \mathbb{R}^{N_T\times N_T}$, and $\mathbf{K}_m^{S,T}\in \mathbb{R}^{N_S\times N_T}$ are the kernel matrices defined for the source domain, the target domain, and the cross domain from the source domain to the target domain, respectively. The above formulation indicates that MMD serve as a good measure of distribution discrepancy in kernel space. For more details about MMD, readers can refer to Borgwardt \cite{MMD}.

With the discrepancy measurement MMD, we can now present the final formulation for DA based on MKL as follows:
\begin{equation}
\label{allobj}
\min\limits_\mathbf{d} \ \frac{1}{2}\Omega^2(\mathbf{d}) + \lambda G(\mathbf{d})
\end{equation}
where $\lambda$ is positive parameters that controls the balance between the domain adaptation and the classification accuracy. Such a formulation actually forces the learnt classifier can simultaneously predict the labels accurately and preserve the semantic relations across domains.

{By introducing domain adaptation, we can explore the available knowledge on a given source domain to develop a classifier built on the target domain where a priori information is not available. This will significantly reduce the heavy requirement of labeled samples in the target domain. Subsequently, our work can enjoy the capability of transferring the model to different target domains easily.}

\subsection{Alternating Optimization}\label{sec:opt}
There are two variables involved in the above formulation, \ie, the variable $\mathbf{d}$ for the MKL in domain adaptation, and variable $\boldsymbol{\alpha}$ for the classifier. We employ the reduced gradient descent procedure proposed in \cite{simplemkl} to iteratively update the linear combination coefficient $\mathbf{d}$ and the dual variable $\boldsymbol{\alpha}$. The optimization consists of two main steps, each of which can be easily solved using the simple and efficient existing techniques.

\begin{itemize}
\item $\boldsymbol{\alpha}$-step: with the fixed $\mathbf{d}$, Problem \ref{allobj} becomes the standard SVM problem defined in Equation \ref{svm}, where $\sum_{m=1}^{M}d_m \mathbf{K}_m^{L,L}$ can be treated as one kernel matrix. We adopted the LIBSVM tool to directly get the optimal $\boldsymbol{\alpha}$.

\item $\mathbf{d}$-step: with the learnt $\boldsymbol{\alpha}$, we can rewrite the problem \ref{allobj} with respect to $\mathbf{d}$ as follows
\begin{equation}\label{opt2}
\begin{split}
\min\limits_{\mathbf{d},\boldsymbol{\alpha}}\ &\mathbf{d}'\mathbf{k}\mathbf{k}'\mathbf{d}-\lambda \mathbf{p}'\mathbf{d}\\
s.t.\  &\mathbf{d}'\mathbf{1}=1, \ {d}_m\geq\mathbf{0}
\end{split}
\end{equation}
with $\mathbf{p} = \left[(\boldsymbol{\alpha}\circ\mathbf{y})'\mathbf{K}^{L,L}_1(\boldsymbol{\alpha}\circ\mathbf{y}),\ldots, (\boldsymbol{\alpha}\circ\mathbf{y})'\mathbf{K}^{L,L}_M(\boldsymbol{\alpha}\circ\mathbf{y})\right]'$. Since the above problem is convex with respect to $\mathbf{d}$, we can directly apply the second-order gradient descent method to solving it.
\end{itemize}

{To obtain the global optimization solution, we alternatingly update the linear coefficient $\mathbf{d}$ and the dual variable $\bm{\alpha}$ in a few iterations. After we have these parameters, the final classier based on the multiple kernels can be formulated as follows
\begin{equation}\label{classfier}
 f(\mathbf{x})+b = \sum_{\mathbf{x}_j\in \mathcal{D}_L} \alpha_j y_j \sum_{m=1}^M d_m k_m(\mathbf{x}_j, \mathbf{x})+b.
\end{equation}}

\subsection{Active Learning}\label{sec:act}
The domain adaption based on MKL can help fully utilize the training data from different domains to obtain a discriminative classifier for HSI. However, as we mentioned above, these exist more or less divergence between the source and the target domains, which limits the power of the domain adaptation. To maximally exploit the semantic information from the target domain and meanwhile minimally rely on the large number of the labeled data, active learning serves as an promising solution to improving the classification performance with a small set of the selected labeled samples \cite{Tong:2002,Jain:2010,Liu:2016:cvpr,Samat:2016}.

{Specifically, in our proposed framework we repeat the training and active learning stages alternatingly, where the active learning stage selects the most informative samples from the target domain, labels it and adds it into the training set $\mathcal{D}_L$, and then the training stage retrain the MKL classifier using the updated dataset $\mathcal{D}_L$ containing more labeled samples from the source domain.}

In the active learning, the selection criteria is quite important for the classification performance. In our framework we adopt the SVM model for HSI classification, which basically pursues the hyperplane that gives the maximum margin. Therefore, in a one-against-all setting for multi-class problems \cite{devis2}, a margin sampling (MS) strategy is employed to heuristically select the best points, \ie, the closest $q$ points $\mathcal{D}_{MS}$, to the hyperplane of the classifier learnt in the last iteration) from the remaining unlabeled dataset $\mathcal{D}_T/\mathcal{D}_C$ according to the following ranking criterion for each candidate $\mathbf{x}_i\in \mathcal{D}_T/\mathcal{D}_C$:
\begin{equation}\label{dist}
h(\mathbf{x}_i) = \left|\sum_{\mathbf{x}_j\in \mathcal{D}_L} \alpha_j y_j \sum_{m=1}^M d_m k_m(\mathbf{x}_j, \mathbf{x}_i)+b\right|.
\end{equation}
Here $h(\mathbf{x}_i)$ is the distance of the samples to the hyperplane defined for any class, with its dual variables $\alpha_i$ and the training data in $\mathcal{D}_L$. Note that when $N_T$ is quite huge, \ie, there are a huge number of the unlabeled samples in the target domain, the above ranking solution will be quite time-consuming due to the expensive computation of the distances for all samples. In this case, we can employ the existing speedup techniques like the popular locality sensitive hashing \cite{Jain:2010,Liu:2016:cvpr}.

Algorithm \ref{MKL-MS} lists the detailed procedures of our proposed active MKL-MS framework.

\renewcommand{\algorithmicrequire}{\textbf{Input:}}
\renewcommand{\algorithmicensure}{\textbf{Repeat:}}
\renewcommand{\algorithmicreturn}{\textbf{Until:}}
\begin{algorithm}[!h]
\caption{Framework of the proposed active MKL-MS}
\label{MKL-MS}
\begin{algorithmic}[1]

\REQUIRE
\STATE Source domain $\mathcal{D}_S = (\mathbf{x}_i^S,y_j)|_{i=1}^{N_S}$, Target domain $\mathcal{D}_T = \mathbf{x}_i^T|_{i=1}^{N_T}$;\\
\STATE Initialize candidate set $\mathcal{D}_C = \varnothing$, the labeled set $\mathcal{D}_L = U_S$, and the weight $d_m = \frac{1}{M}$;\\

\ENSURE
\STATE Calculate the base kernels $\mathbf{K}_m|_{m=1}^M$ according to Equation (\ref{kernel});\\
\STATE Train the multi-kernel classifier using $\mathcal{D}_L$ and the corresponding labels by solving Problem (\ref{svm}) using LIBSVM;\\
\STATE Optimize the multi-kernel combination by solving Problem (\ref{opt2}) using gradient descent method;
\STATE Score the unlabeled samples $\mathbf{x}_i\in \mathcal{D}_T/\mathcal{D}_C$ according to the distance $h(\mathbf{x}_i)$ in Equation \ref{dist};\\
\STATE Select the top-$q$ closest samples $\mathcal{D}_{MS}$ from $\mathcal{D}_T/\mathcal{D}_C$;\\
\STATE Annotate $\mathbf{x}^{MS}$ and update the candidate set $\mathcal{D}_C = \mathcal{D}_C \cup \mathcal{D}_{MS}$;\\
\RETURN
The stop criterion is satisfied.
\end{algorithmic}
\end{algorithm}

\section{Experiments}\label{sec:exp}

\begin{figure}[!t]
	\centering
	\subfigure[Pavia Center]{
		\begin{minipage}[t]{0.4\linewidth}
			\centering
			\includegraphics[width=0.45\textwidth]{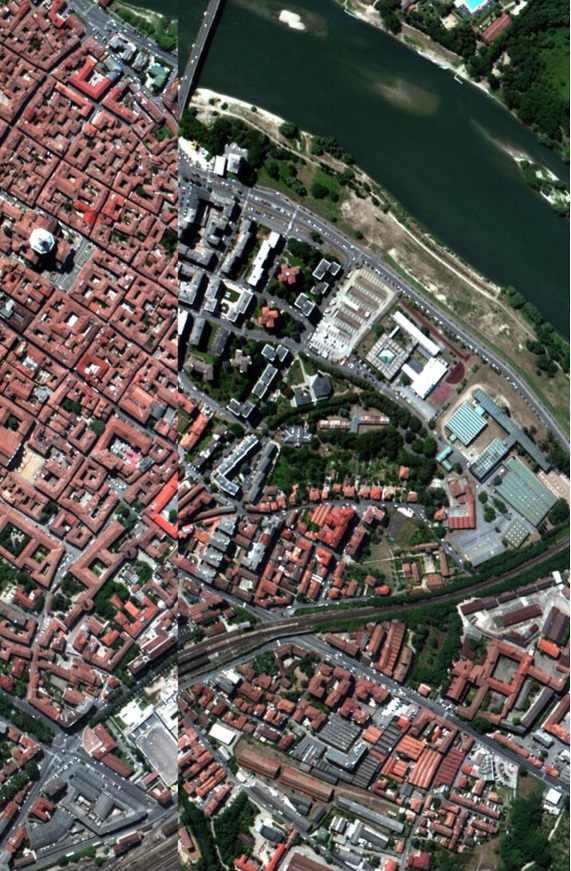}
			\label{pavia-fig-1} %
			\hspace{5pt}
			\includegraphics[width=0.45\textwidth]{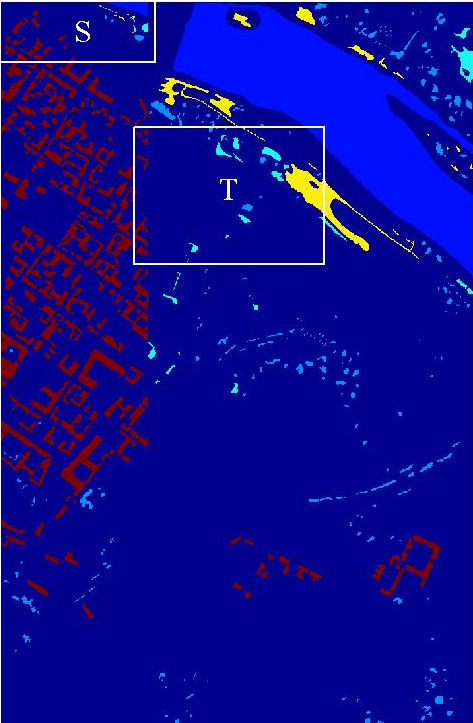}
			\label{pavia-fig-2}
	\end{minipage}}
	\vspace{0.1pt}
	\subfigure[University Area]{
		\begin{minipage}[t]{0.4\linewidth}
			\centering
			\includegraphics[width=0.383\textwidth]{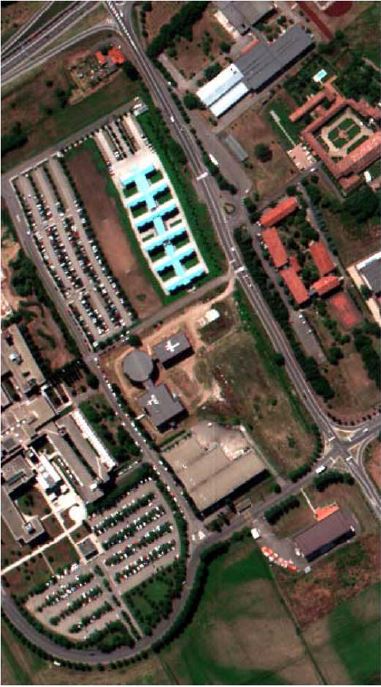}
			\label{paviau-fig-1}
			\hspace{5pt}
			\includegraphics[width=0.383\textwidth]{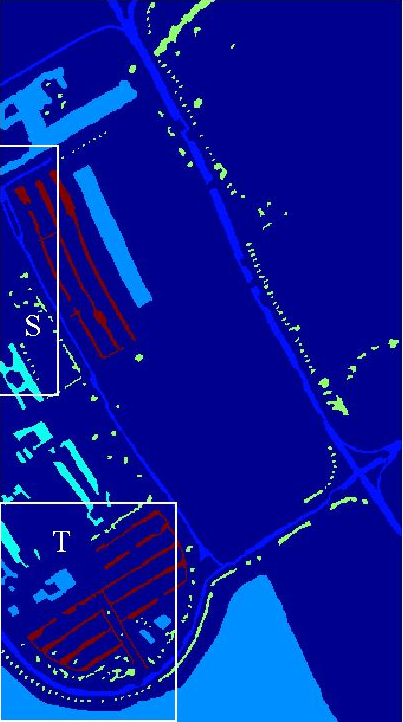}
			\label{paviau-fig-2}
	\end{minipage}}
	\caption{The left column is false color composition and the right one is ground-truth map containing five different land-cover classes: (a) Pavia Center and (b) University Area}
\end{figure}

\begin{figure}[!t]
\centering
\subfigure[Pavia Center]{
\includegraphics[width=0.375\textwidth]{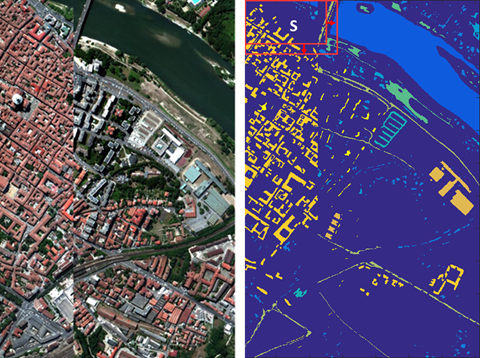}
}\hspace{0.15in}
\subfigure[University Area]{
\includegraphics[width=0.35\textwidth]{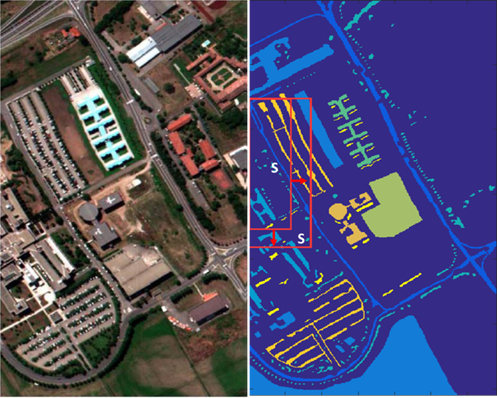}
}\label{paviau-fig-large}
\caption{The left column is false color composition and the right one is ground-truth map with the enlarged source regions: (a) Pavia Center and (b) University Area}
\end{figure}

\subsection{Settings and Protocol}
In the experiments, we employ two hyperspectral datasets to evaluate the proposed method, which adapts the multi-kernel classifier trained on the area S (considered as source domain) to the spatially separate area T (considered as target domain). The first dataset is Pavia center (northern Italy) containing $1096\times1096$ pixels. As shown in Fig. \ref{pavia-fig-2}, the ground-truth map with five classes (water, trees, meadow, soil and tile) of interest available for the scene, displayed in the form of a class assignment for each labeled pixel. These classes have been included in a labeled data set of 126,069 samples extracted by visual inspection. The second hyperspectral dataset is the university area, whose image size is $610\times340$ in pixels. Similar to the Pavia center, it is also divided into nine classes. Fig. \ref{paviau-fig-2} presents the five reference classes of interest, i.e., asphalt, meadows, gravel, trees, and bricks. Totally, the number of different labeled samples available for university area scene is 34,125.

On each dataset, 20 samples of each class (i.e., 100 samples totally for five classes) were randomly selected as the initial training set to obtain a multi-kernel classifier. To suppress the randomness in the evaluation, all the results are averaged over ten times of experiments, namely, we sample ten different initial training sets from the source domain. The active learning process runs with two settings: selecting and adding ${q} = 10$ and ${q} = 20$ pixels into the training set per iteration. For the above settings, we respectively repeat 40 and 30 iterations for Pavia center image, and 30 for University area in all cases.

{As to the multi-kernel classifier, we following the common way for multi-class classification problem that one-versus-all classifiers are trained with four types of kernels (\ie, $M=4$): Gaussian kernel $k_1(\mathbf{x}_i,\mathbf{x}_j)=\exp(-\gamma\|\mathbf{x}_i-\mathbf{x}_j\|^2)$, Laplacian kernel $k_2(\mathbf{x}_i,\mathbf{x}_j)=\exp(-\sqrt{\gamma}\|\mathbf{x}_i-\mathbf{x}_j\|)$, inverse square distance kernel $k_3(\mathbf{x}_i,\mathbf{x}_j)=\exp(\frac{1}{\gamma\|\mathbf{x}_i- \mathbf{x}_j\|^2+1})$, and inverse distance kernel $k_4(\mathbf{x}_i,\mathbf{x}_j)=\exp(\frac{1}{\sqrt{\gamma}\|\mathbf{x}_i-\mathbf{x}_j\|+1})$. Here $\gamma$ represents the kernel parameter, for which we use the default value $\gamma = \frac{1}{dist^*}$, with respect to the mean value $dist^*$ of the square distances between all training samples.}

{We compare our method with those classic active learning methods with different heuristics like margin sampling (MS), random sampling (RS), and MKL-RS (with random sampling heuristic), in terms of the classification accuracy (\%) and Kappa statistic. The Kappa statistic is a metric that compares an bbserved accuracy with an expected accuracy (random chance) \cite{Viera:2005}. It takes into account random chance (agreement with a random classifier), which generally means it is a more robust measure than simple percent agreement calculation. We adopted it for the comprehensive evaluation of the proposed method in our paper.} For these active learning methods, a small set of labeled data in source domains with the sequentially selected samples in the target domain are treated as the training data. In addition, we also adopt those traditional methods without active learning process (non-AL for short), like single kernel version of DTMKL (SKV), classic SVM, and domain transfer multi-kernel learning (DTMKL) \cite{zhuo}, as the baselines, where they utilize all the labeled samples in source domains to train the classifier for the target domain. In those kernel methods we simply choose the Gaussian kernel the default one. As to the SVM based methods, the popular LIBSVM is applied in the experiments, and the optimal SVM parameters are found using the grid search in a tenfold cross-validation manner, where the parameter ${C} \in\{2^{-1},\ 2^{0},\ 2^{1},\ 2^{2}\}$ and $\lambda\in\{2^{-4},\ 2^{-3},\ 2^{-2},\ 2^{-1}\}$.

\begin{table*}[!hpt]
\caption{Classification Accuracy (\%) with Respective to the Number of Samples Added to Target Domain on Pavia Center}
\label{pavia-table-3}
\centering
\begin{tabular}{c|c|c|c|c}
\hline
\multirow{2}{*}{Method} & \multicolumn{4}{c}{\#  target samples ($q=20$)}\\
\cline{2-5}
 & 0 & 40 & 100 &  200\\
\hline
RS & 47.770 & 86.200 & 90.445 & 93.865\\
\hline
MS & 47.770  & 87.405 & 95.685 & 96.255\\
\hline
MKL-RS & 47.930  & 92.250 & 94.275 & 94.655\\
\hline
MKL-MS & \textbf{47.945}  & \textbf{94.765} & \textbf{98.610} & \textbf{98.235}\\
\hline
\end{tabular}
\end{table*}

\begin{table*}[!t]
\caption{Classification Accuracy (\%) and Kappa Statistic on the Pavia Center Dataset}
\label{pavia-table-2}
\centering
\begin{tabular}{c|c|c|c}
\hline
Method &  \# source/target samples & OA  & Kappa\\
\hline
SKV & 2723  /  0  & 46.550 & 0.358 \\
\hline
DTMKL  & 2723 /  0   & 46.850 & 0.360 \\
\hline
SVM &   2723 /  0   & 49.750 & 0.388 \\
\hline
RS & 100  / 300 & 92.220 $\pm$ 5.288 & 0.882 $\pm$ 0.085\\
\hline
MS & 100  / 300 & 96.140  $\pm$ 2.427 & 0.943  $\pm$ 0.036\\
\hline
MKL-RS &  100  / 300 & 92.960 $\pm$ 2.383 & 0.897  $\pm$ 0.035 \\
\hline
MKL-MS &  100 / 300 & \textbf{97.930}  $\pm$ {0.658} & \textbf{0.970}  $\pm${0.010}\\
\hline
\end{tabular}
\end{table*}

\begin{table*}[!t]
\caption{{Classification Accuracy (\%) and Kappa Statistic on the Pavia Center Dataset with different size of source region}}
\label{pavia-table-3}
\centering
\begin{tabular}{c|c|c|c|c}
\hline
region size & Method & \# source/target samples & OA  & Kappa\\
\hline
\multirow{4}{*}{\rotatebox{0}{{3842}}}
    & \multirow{2}{*}{\rotatebox{0}{\scriptsize{MKL-RS}}}
        & 80 / 300	& 93.244$\pm$0.215	& 0.914$\pm$0.008\\
    &    & 100/300	& 92.825$\pm$0.153	& 0.922$\pm$0.013\\\cline{2-5}

    & \multirow{2}{*}{\rotatebox{0}{\scriptsize{MKL-MS}}}
        & 80/300	& 98.301$\pm$0.322	& 0.975$\pm$0.010\\
    &    & 100/300	& 97.467$\pm$0.235	& 0.963$\pm$0.002\\\hline

\multirow{4}{*}{\rotatebox{0}{{5338}}}
    & \multirow{2}{*}{\rotatebox{0}{\scriptsize{MKL-RS}}}
        & 80 / 300	& 90.132$\pm$0.056	& 0.903$\pm$0.006\\
    &    & 100/300	& 90.098$\pm$0.028	& 0.892$\pm$0.016\\\cline{2-5}

    & \multirow{2}{*}{\rotatebox{0}{\scriptsize{MKL-MS}}}
        & 80/300	& 94.200$\pm$0.032	& 0.923$\pm$0.005\\
    &    & 100/300	& 94.067$\pm$0.005	& 0.912$\pm$0.016\\\hline
\end{tabular}
\end{table*}

\subsection{Results and Analysis}

\subsubsection{Experiments on Pavia Center Dataset}
\begin{figure*}[!t]
\centering
\subfigure[${q}=10$ on Pavia center]{
	\centering
	\includegraphics[width=0.475\textwidth]{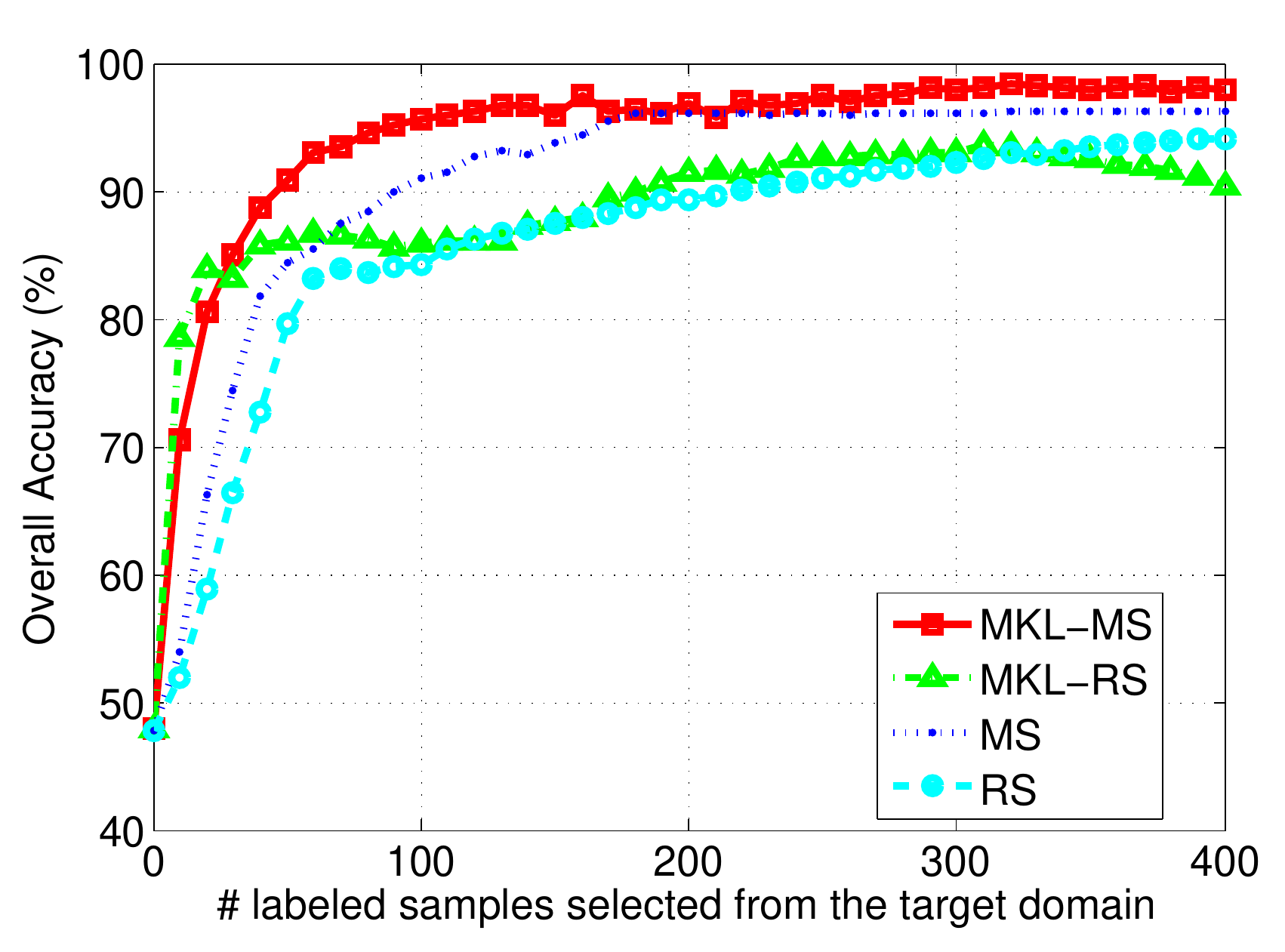}	\label{pavia-fig-3}
}\hspace{0.0in}
\subfigure[${q}=20$ on Pavia center]{
	\centering
	\includegraphics[width=0.475\textwidth]{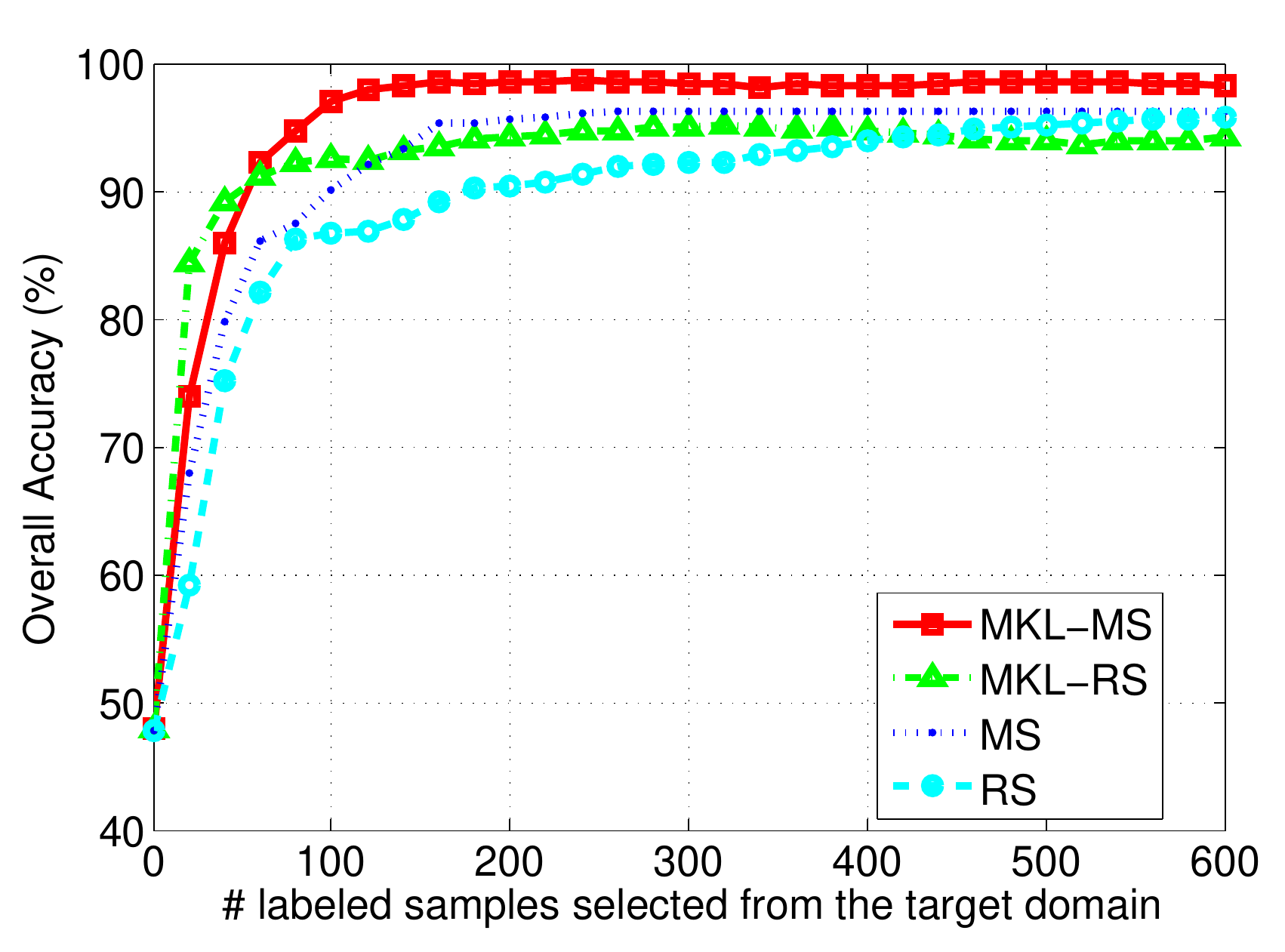}
	\label{paviau-fig-3}
}\\

\subfigure[${q}=10$ on University Area]{
	\includegraphics[width=0.475\textwidth]{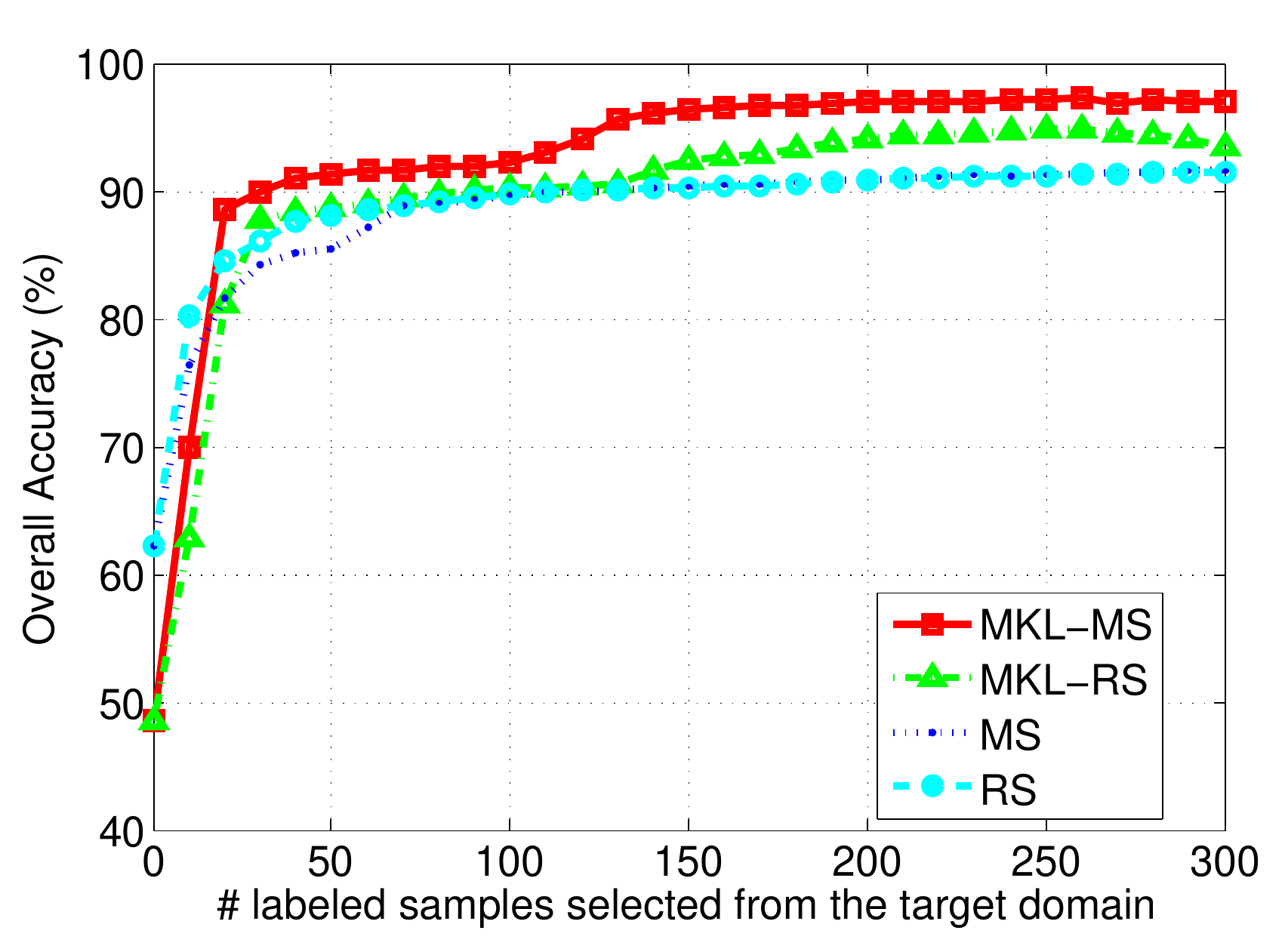}
	\label{pavia-fig-4}
}\hspace{0in}
\subfigure[${q}=20$ on University Area]{
	\includegraphics[width=0.475\textwidth]{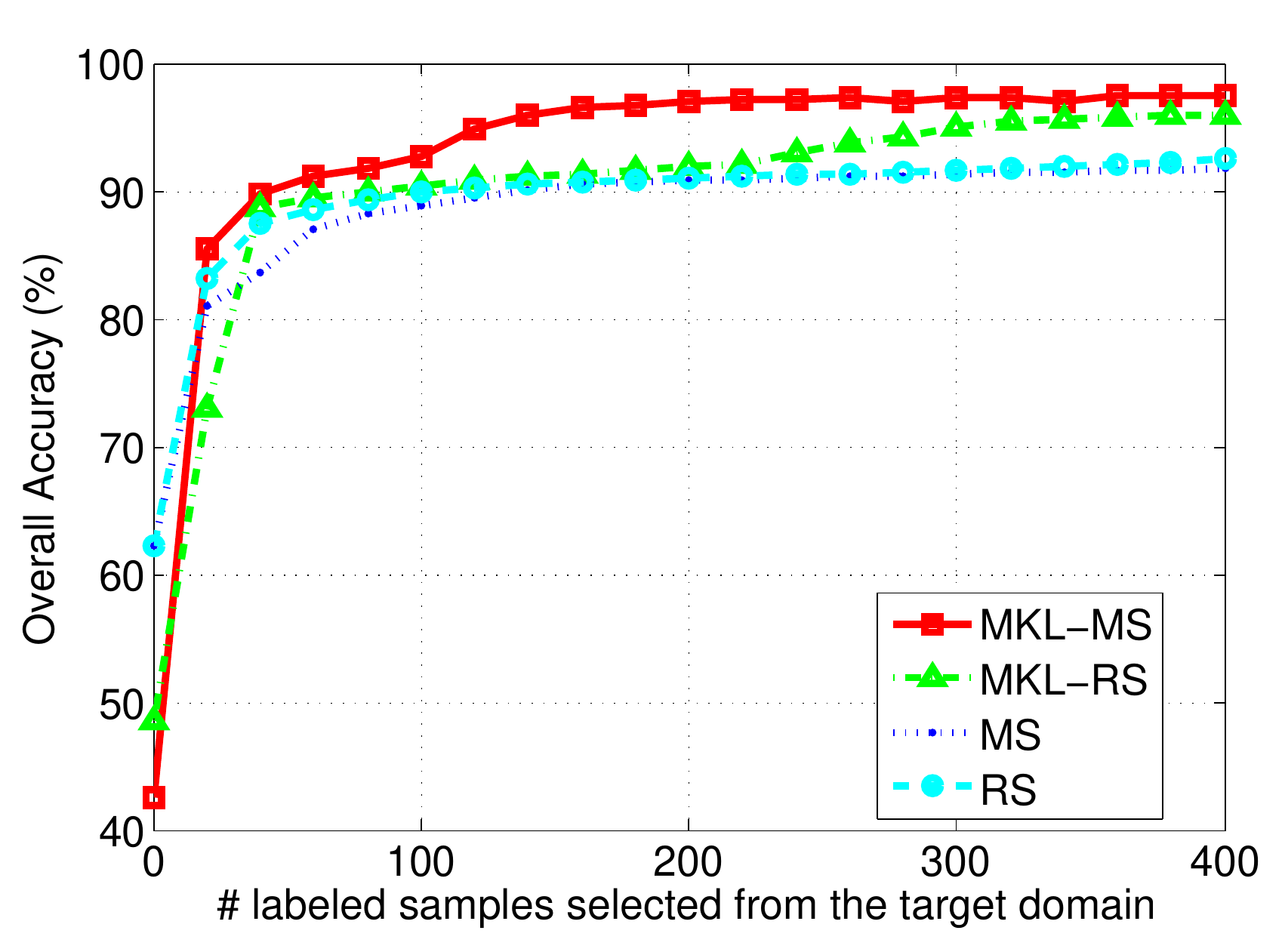}
	\label{paviau-fig-4}
}
\caption{The learning curves on Pavia center and University Area, where RS (cyan line) = random sampling, MS (blue line) = margin sampling, MKL-RS (green line) = random sampling combined with MKL, and MKL-MS (red line) = margin sampling combined with MKL. ${q}$ presents the number of pixels selected from the target domain.}
\label{learning curve1}
\end{figure*}

Fig. \ref{pavia-fig-3} plots the obtained overall accuracy (OA) of the classification using four different AL methods: RS, MS, MKL-RS, and MKL-MS, with respect to the number of labeled training samples for the Pavia center scene. From the figure, we can first notice that when the training set is very small, the proposed MKL-MS provides lower OA than MKL-RS. But as the number of training samples increases, e.g., to 140 at the fourth iteration, MKL-MS substantially improves its classification accuracy, and gets the best performance among all methods. Besides, we can also notice that MKL-MS attains a quick convergence, after about 10 iteration with 100 target domain pixels selected, and its OA curve (in red) reaches 97.4\% (about 4.6\% performance gains over MKL-RS). This performance is never reached by the three baseline approaches during the 40 active learning iterations.

Table \ref{pavia-table-3} further investigates the OA for different numbers of samples added during the AL process. As it could be expected, more added samples indicate a higher classification accuracy. Moreover, the proposed MKL-MS better behaves compared to other approaches with a obvious performance gap. Table \ref{pavia-table-2} reports the results of three non-AL methods and four active learning based methods, in terms of OA, Kappa, and their standard deviations. Here, those non-AL algorithms like SKV, classic SVM and DTMKL utilize the whole source domain (i.e., 2723 samples in Pavia Center) as the training set to adapt the learnt model for target domain. {We can we can notice that even though with much more training samples, the classification accuracy of non-AL methods still can not outperform AL strategies with DA technique, which even can not reach 50\% of OA in Table \ref{pavia-table-2}. This is because that there exists huge distribution variation between the source and the target domains}. Instead, our proposed method MKL-MS incorporating the active learning to select the most informative samples to reduce their distribution discrepancy, and thus reaches 97.4\% OA with totally 400 samples, including 100 samples from source domain and 300 samples from target domain. The observation clearly verifies the effectiveness of our strategy combining MKL and AL.

{To comprehensive evaluate the performance, we randomly select more and different source regions from the two datasets and keep the same target image as the prior experiments. The performance further confirm that our proposed method with MS strategy (MKL-MS) again obtain the best performance. This means that there exist the obvious domain shift, and our method can capture the correlations, select the most informative samples for a better classifier, and consistently achieve the best performance. Besides, we also compare our method with the state-of-the-art classification methods including SKV, DTMKL, and SVM. With the help the samples from target domains, RS, MS, MKL-RS, MKL-MS can significantly improve the performance. Besides, by considering the domain shift, MKL-MS method can obtain the best performance in most cases. This further confirm that our method can not only largely leverage the information from target domain, but also exploit the information from source domains, and thus show the robust and better performance in practice.}

{To further investigate the effect of the source region size, we gradually enlarge the source region from 3842 samples to 5338 ones on Pavia Center, and keep the target region unchanged. Fig. \ref{paviau-fig-large}) demonstrate the region growth, and Table \ref{pavia-table-3} lists the performance using different number of labeled samples in our framework, where we can see that MKL-MS outperforms MKL-RS consistently, and get the best performance compared to the other baselines (in Table \ref{pavia-table-2}) in all cases.}

\begin{table*}[!t]
\caption{ Classification Accuracy  (\%) with Respective to the Number of Samples Added to Target Domain on University Area}
\label{paviau-table-1}
\centering
\begin{tabular}{c|c|c|c|c}
\hline
\multirow{2}{*}{Method} & \multicolumn{4}{c}{\# target samples (q=20)}\\
\cline{2-5}
 & 0 & 40 & 100 &  200\\
\hline
RS & 62.290 & 87.385 & 89.955& 91.030\\
\hline
MS & \textbf{62.290}  & 83.600 & 88.910 & 90.785\\
\hline
MKL-RS & 48.575  & 88.625 & 90.320 & 91.850\\
\hline
MKL-MS & 42.525  & \textbf{89.830} & \textbf{92.625} & \textbf{96.925}\\
\hline
\end{tabular}
\end{table*}

\begin{table*}[!t]

\caption{Classification Accuracy (\%) and Kappa Statistic on the University Area Dataset}
\label{paviau-tabel-2}
\centering
\begin{tabular}{c|c|c|c}
\hline
Method & {\# source/target samples} & OA  & Kappa \\
\hline
SKV &1907 / 0 & 48.350 & 0.3385\\
\hline
DTMKL & 1907 / 0 & 44.800 & 0.298\\
\hline
SVM & 1907 / 0 & 64.650  & 0.526 \\
\hline
RS & 100 / 300 & 91.430 $\pm$ 0.4264 & 0.871 $\pm$ 0.006\\
\hline
MS & 100 / 300 & 91.610 $\pm$ 0.5592 & 0.873 $\pm$ 0.009\\
\hline
MKL-RS & 100 / 300 & 93.480 $\pm$ 1.214 & 0.901 $\pm$ 0.019\\
\hline
MKL-MS & 100 / 300 & \textbf{97.070} $\pm$ {0.343} & \textbf{0.956} $\pm$ {0.005}\\
\hline
\end{tabular}
\end{table*}

\begin{table*}[!t]
\caption{{Classification Accuracy (\%) and Kappa Statistic on the University Area Dataset with different size of source region}}
\label{paviau-table-3}
\centering
\begin{tabular}{c|c|c|c|c}
\hline
region size & Method & \# source/target samples & OA  & Kappa\\
\hline

\multirow{4}{*}{\rotatebox{0}{{2842}}}
    & \multirow{2}{*}{\rotatebox{0}{\scriptsize{MKL-RS}}}
        & 100/300	& 94.825$\pm$0.825	& 0.922$\pm$0.013\\
    &    & 200/300	& 94.553$\pm$0.157	& 0.918$\pm$0.018\\\cline{2-5}

    & \multirow{2}{*}{\rotatebox{0}{\scriptsize{MKL-MS}}}
        & 100/300	& 97.250$\pm$0.235	& 0.956$\pm$0.003\\
    &    & 200/300	& 96.950$\pm$0.453	& 0.952$\pm$0.004\\\hline

\multirow{4}{*}{\rotatebox{0}{{4623}}}
    & \multirow{2}{*}{\rotatebox{0}{\scriptsize{MKL-RS}}}
        & 100/300	& 93.550$\pm$1.55  &	0.902$\pm$0.024\\
    &    & 200/300	& 93.950$\pm$0.05 &	0.909$\pm$0.001\\\cline{2-5}

    & \multirow{2}{*}{\rotatebox{0}{\scriptsize{MKL-MS}}}
        & 100/300 & 96.879$\pm$0.225	&¡¡0.953$\pm$0.004\\
    &    & 200/300 & 96.825$\pm$0.225	& 0.952$\pm$0.003\\\hline
\end{tabular}
\end{table*}

\subsubsection{Experiments on University Area Dataset}
To comprehensively evaluate the proposed method, in Fig. \ref{paviau-fig-3} we further display the learning curves obtained on University Area dataset using the four active learning methods. The proposed MKL-MS algorithm clearly obtains the best results, which certainly demonstrates the advantage of MKL-MS. Similar to the results on Pavia center, at the beginning the proposed MKL-MS lies 15\% below the traditional method (MS and RS) without adaptation. However, after three active iterations MKL-MS algorithm reaches nearly 90\% OA. At the same time, it can be observed that MKL-RS also archives a promising learning curve, which is closest to that of MKL-MS among all methods. This fact confirms that selecting samples near the decision boundary helps boost the classification performance, and significantly outperforms the random sampling. In this figure, all methods converge quickly after adding 150 samples from the target domain, where we can see that using $q=20$ we can achieve much more efficiency with fewer active iterations than $q=10$. Moreover, in all cases our proposed method MKL-MS achieves the best performance, i.e., nearly 97.0\% OA, while only about 95.0\% using  MKL-RS, 91.5\% using MS and RS.

In Table \ref{paviau-table-1}, we also list the classification accuracy of the four methods by varying the number (ranging from 0 to 200) of samples selected from target domain. From the table, we can see that MKL-MS always sustains the highest OA from the beginning of the AL process. Table \ref{paviau-tabel-2} further reports the OA and Kappa results of three non-AL and four AL classification methods. For these non-AL algorithms, we still use all the source domain (i.e., 1907 samples in University Area) as training set to train the classifier. {We can see that there exists significant domain shift between the source and target domains, by comparing the basic classification methods and the active learning methods with DA. Without considering the divergence between the source and target domains, all the non-AL methods that directly transfer the domain knowledge, get unsatisfying performance in both OA and Kappa. Owing to the AL strategies, our proposed approach MKL-MS only relies on a very few training samples and obtains promising results.} For example, in Fig. \ref{paviau-fig-3} with $q=10$ it takes only 15 iterations (150 samples are labeled and added into training set) to converge at the level of 96.9\% accuracy, and while with $q=20$, 140 samples are selected totally. By comparing its performance to the three non-AL classification methods, we can conclude that the active learning strategy would be very helpful for discriminative classifier learning, when there exists a large distribution deviation from the source domain to the target domain. {In Table \ref{paviau-table-3} we also investigate the performance with respect to different source region size, and obtain the similar conclusion that our proposed method is able to robustly achieve the promising performance in different scenarios.}

\section{Conclusion}\label{sec:con}
This paper presents a novel active framework MKL-AL based on domain adaptation for addressing the hyperspectral image (HSI) classification problem. It fully utilizes the label information from the auxiliary domains by compensating domain distribution shifts in an sequential active learning manner, which not only significantly boosts the classification accuracy but also saves the expensive computational cost and human labelling efforts. The extensive experimental results on two popular HSI datasets demonstrated that when a large bias exists between the source and target domains, the conventional DA classifier can not promise a satisfying performance. Instead, by actively expanding the training set without too much efforts, our method can efficiently improve the classification accuracy.

\section{Acknowledgement}
This work was supported by the National Natural Science Foundation of China (61402026 and 61572388), the Key R\&D Program - The Key Industry Innovation Chain of Shaanxi (Grant No. 2017ZDCXL-GY-05-04-02 and Grant No. 2017ZDCXL-GY-05-02), Beijing Municipal Science and Technology Commission (Z171100000117022) and the Foundation of State Key Lab of Software Development Environment (SKLSDE-2016ZX-04).

\section*{References}

\bibliography{elsarticle-template-cr}

\end{document}